\pdfoutput=1

\documentclass[11pt]{article}

\usepackage{acl}

\usepackage{times}
\usepackage{latexsym}
\usepackage{graphicx}

\usepackage[T1]{fontenc}

\usepackage[utf8]{inputenc}

\usepackage{microtype}

\title{Emotion-Cause Pair Extraction in Customer Reviews}

\author{Aishwarya Kaliki \\ \texttt{akaliki@usc.edu} \And Arpit Mittal \\ \texttt{arpitm@usc.edu} \AND Jeel Tejaskumar Vaishnav \\ \texttt{jvaishna@usc.edu} \And Nathan Johns \\ \texttt{johnsn@usc.edu} \And Wyatt Pease \\ \texttt{wlpease@usc.edu}}

\begin{document}
\maketitle

\begin{abstract}
Emotion-Cause Pair Extraction (ECPE) is a complex yet popular area in Natural Language Processing due to its importance and potential applications in various domains. In this report , we aim to present our work in ECPE in the domain of online reviews. With a manually annotated dataset, we explore an algorithm to extract emotion cause pairs using a neural network. In addition, we propose a model using previous reference materials and combining emotion-cause pair extraction with research in the domain of emotion-aware word embeddings,  where we send these embeddings into a Bi-LSTM layer  which gives us the emotionally relevant clauses. With the constraint of a limited dataset, we achieved . The overall scope of our report comprises of a comprehensive literature review, implementation of referenced methods for dataset construction and initial model training, and modifying previous work in ECPE by proposing an improvement to the pipeline, as well as algorithm development and implementation for the specific domain of reviews. 
\end{abstract}

\section{Introduction}

Emotions play an important role in the decision-making process of humans and thus, in recent years, extracting emotion from text has seen increased interest due to its possible application in a large number of domains such as business, politics, psychology and social media. However, identifying and extracting emotion is a very complicated task due to the innate subtlety and complexity of emotional expressions and language. It is also a multiclass classification problem that combines both core machine learning techniques in addition to natural language. In many cases, there is also a need to identify the cause of the observed emotions as well, and this has given rise to research on emotion-cause extraction techniques. In their study, \citet{9478079} reviewed the most recent research in this field, detailing the corpus, methodology and evaluation. 
 
An interesting application of emotion-cause extraction is its importance in customer reviews \citet{gupta-etal-2010-emotion}. From household items, books or movies to hotels and restaurants, reviews are a great way for a consumer to familiarize themselves with the merits and drawbacks of said commodity. However, in many instances, the volume of reviews is so high that it would be impractical to read them all. Emotion-cause extraction would help in understanding why the emotion expressed in the review came to be, and can help pinpoint key issues.

In this project, we use Emotion-Cause Extraction along with Clustering techniques to provide an overview of a product based on the reviews and star-ratings. Firstly, we find the Word2Vec word embeddings for the words in the reviews, and enhance them with emotional context. (Mao et al.) Next, we use clause extraction, a RNN emotion extraction model and a RNN cause extraction model to get cause clause for each review. Then, we vectorize the cause clauses and prepare the data for Agglomerative clustering, clustering reviews for each (product, emotion) pair based on its vector representation. As the final step, for each cluster, we find a head clause and provide a list of the most important or prevalent causes for each (product, emotion) pair. 

The goal of this project was to provide a nlp-based method to analyze customer reviews and extract key issues, thereby enabling manufacturers, retailers and service providers to foster a better understanding of the performance of their product or improve their relationship with their customer. It will also help customers to get a better idea of the product without going through the large number of reviews. We achieve this by combining ECPE techniques with other machine learning techniques to build a pipeline for processing reviews in a way that makes them easy to analyze and understand.

\section {Related Work}

\citet{lee-etal-2010-text} first presented the task of emotion cause extraction (ECE) and defined this task as extracting the word-level causes that lead to the given emotions in text. They constructed a small scale Chinese emotion cause corpus in which the spans of both emotion and cause were annotated. Based on the above task, there were some other studies that conducted ECE research on their own corpus using rule based methods or machine learning methods.

\citet{Chen10emotioncause} suggested that a clause may be the most appropriate unit to detect causes based on the analysis of the corpus in \citet{lee-etal-2010-text} and transformed the task from word-level to clause-level. They proposed a multi-label approach that detects multi-clause causes and captures the long-distance information. There was a lot of work done based on this task setting. \citet{russo-etal-2011-emocause} introduced a method based on the linguistic patterns and common sense knowledge for the identification of Italian sentences which contain a cause phrase. \citet{GuiYXLLZ14} used 25 manually compiled rules as features, and chose machine learning models, such as SVM and CRFs, to detect causes. \citet{gui-etal-2016-event} and \citet{8195347} released a Chinese emotion-cause dataset using SINA city news. This corpus has received much attention and has become a benchmark dataset for ECE research. Based on this corpus, several traditional machine learning methods and deep learning methods were proposed.

All of the above work attempts to extract word-level or clause-level causes given the emotion annotations. \citet{xia-ding-2019-emotion} proposed a two-step process to extract emotion-cause pairs from documents without the need for annotation. This process first performs individual emotion extraction and cause extraction via multi-task learning, and then conducts emotion-cause pairing and filtering. The experimental results for this work on a benchmark emotion cause corpus \citep{gui-etal-2016-event} were quite high. More recently, end-to-end networks such as that proposed by \citet{singh2021endtoend} have shown to provide further improvements over multi-stage approaches by leveraging the mutual dependence between the extracted emotion clauses and cause clauses. \cite{li-etal-2018-co} also proposed a context-aware co-attention model for emotion cause pair extraction. \citet{cheng-etal-2020-symmetric} detailed a methodology that uses local search to pair emotions and causes simultaneously, reducing the search space and improving efficiency.

In the domain of Sentiment-Aware Word Embeddings, \citet{app9071334} details a thorough algorithm to add emotional context to Word2Vec embeddings, as well as shows quantifiable improvement in results when used in emotional classification tasks. 

\section {Architecture and Implementation}

The pipeline implemented in the project to achieve our objective can be split into the following steps: data annotation, creation of emotion aware word embeddings, extraction of clauses, extraction of emotions, extraction of causes of said emotions, and clustering of causes to summarize the reviews. Each of these steps are detailed in the following sections.

\subsection{Data Annotation}

Since there is no benchmark dataset for the ECPE task in the English language and no review dataset annotated with emotions, we have annotated the data manually. We picked 1000 diverse data points equally across 50 products and for each review, we manually annotated the emotion and cause clause pairs. 

\subsection{Creation of Emotion Aware Word Embeddings}

The process for creating emotion-aware word embeddings centered around the idea that words have and use emotional context in addition to syntactic and structural context. To add this to our pipeline, we first start with Word2Vec embeddings for all words in the corpus. This is done using the gensim library and its inbuilt implementation. 

Next, we use an emotion lexicon (in our case we used \url{http://saifmohammad.com/WebPages/lexicons.html}) to identify emotion words in the English language as well as the intensity associated with them. 

To identify the emotional context of a word, we calculate the cosine similarity of that word with every emotion word in the lexicon that also occurs in our vocabulary. Doing this for all words and all emotion words generates a similarity matrix between the vocabulary of our corpus and the emotional words present in it. 

Adding the emotional context is a slightly more complicated task since any one word can be similar to multiple emotion words. One way to deal with this is to impose a limit on the number of emotions to add as context. In our specific use case, we decided to set that limit to 2. This is because in the domain of reviews, it is very rare to see more than two main emotions in any one given document.

For each word in the vocab, we then calculated the average emotion embedding by combining the Word2Vec embeddings of the 2 most similar emotion words as a weighted average. The weight of each individual emotion in the final embedding is a function of both the similarity to the original word as well as the intensity of the emotion expressed. 
Finally, we overlay this emotion embedding with the original Word2Vec embedding of the word to obtain an emotion-aware word-embedding. We repeat this for all words in the vocabulary.

\begin{figure*}  
    \includegraphics[width=\textwidth]{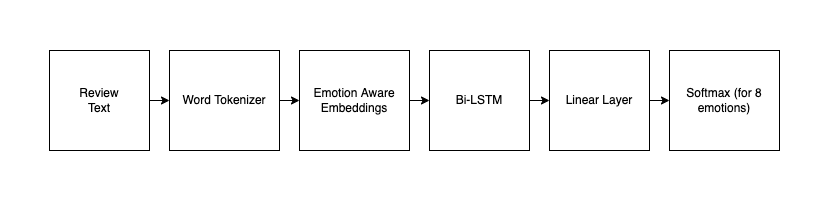}  
    \caption{Model to Extract Emotions}  
    \label{fig:emoExt}
\end{figure*}

\subsection{Extraction of Clauses}

To prepare the reviews for processing, we used nltk inbuilt sentence tokenizer to tokenize the sentences from each review. For clause generation, we used the spaCy NLP package (specifically utilizing the en-web-core-trf module for accuracy) to parse the clauses , using each sentence’s predicted dependency structure. A dependency structure is determined by the relation between a word (a head) and its dependents, wherein each word is connected to each other by directed links or dependency. 

The spaCy engine will determine a ROOT word (usually a finite verb), which forms the structural center of our clause. The roots form the starting point from which the directed links are generated to every other word in the sentence’s clauses. We can now predict the clause by finding dependencies of words present before and after the root. Each sentence was processed using the spacy engine, which created the dependency structure for each sentence as well as predicted the parts of speech for each word. 

To find the ROOT word of the intended clause (as explained above), we call findRootOfSentence function, which returns the token that has a dependency tag of ROOT as well as the predicted dependency structure for the sentence. 

Next, we determine the VERB tokens present in our sentence and their direct dependency on the ROOT word. For this, we used the findOtherVerbs function, to look for tokens that have the VERB part-of-speech tag and has the root token as its only ancestor. In doing this we are finding clausal heads within the sentence that form separate clauses.

Using the getClauseTokenSpanForVerb function, we find the beginning and ending index for the verbs found previously. The function goes through all the verb's children; the leftmost child's index is the beginning index, while the rightmost child's index is the ending index for this verb's clause. 

Next we found the clause indices for each verb. The tokenSpans variable contains the list of tuples, where the first tuple element is the beginning clause index and the second tuple element is the ending clause index.

Finally, we create the token Span objects (using the above tokenSpan variables for each clause) in the sentence. We get the Span object by slicing the Doc object and then appending the resulting Span objects to a list. 

As a final step, we sort the list to make sure that the clauses in the list are in the same order as in the sentence. The final result is a list of documents tokenized into sentences, which are further split into clauses and then words within.

\subsection{Extraction of Emotions}

\begin{figure*}
    \centering  
    \includegraphics[width=\textwidth]{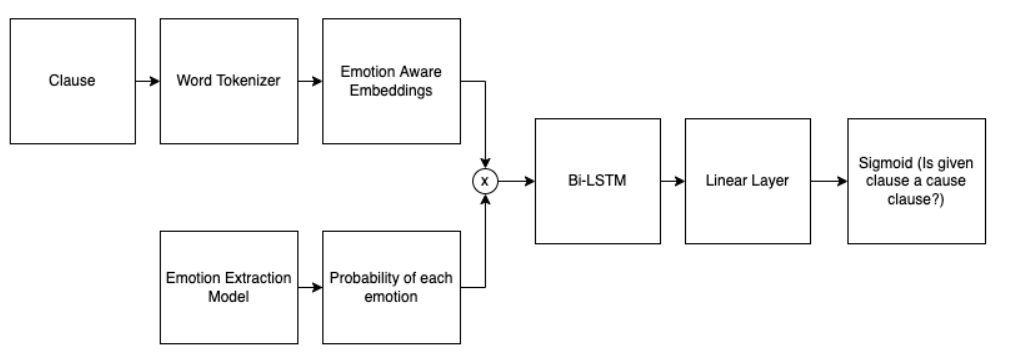}  
    \caption{Model to Extract Causes}  
    \label{fig:causeExt}
\end{figure*}

For emotion extraction, we are using review text as input. Firstly, we tokenize the review text into words. Then we convert each word into their emotion-aware embedding. These emotion aware embeddings are then passed into an RNN-classifier (consisting of a BRNN followed by a classifier) which classifies the review text into 8 emotion classes. The output of this model will be used as a signal to the cause extraction model.

The word embeddings are used as a time-step input to the Bi-LSTM. The Bi-LSTM outputs a 2 * 256 size vector for each time-step. The output of the last time step of Bi-LSTM is passed through a dropout with probability 0.5. We use the output of the last layer as input to a linear layer. The linear layer outputs a vector of size 80 which is passed through the elu nonlinearity function. The output of this layer is further passed to a linear layer which outputs a vector of size 8. This output is further passed to a log softmax function which provides the log-probability for each of the 8 emotions. The training of the model is carried out with the help of the NLL loss function. We are running a stochastic-gradient descent and hence, are using a batch size of 1. 

\begin{figure*}  
    \includegraphics[width=\textwidth]{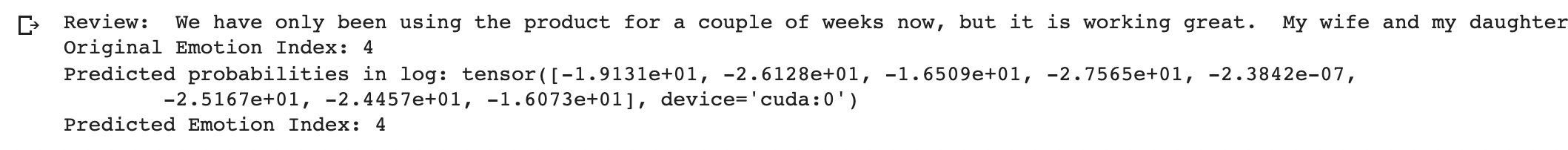}  
    \caption{Results of Emotion Extraction on Sample 1}  
    \label{fig:resultsE1}
\end{figure*}

\begin{figure*}  
    \includegraphics[width=\textwidth]{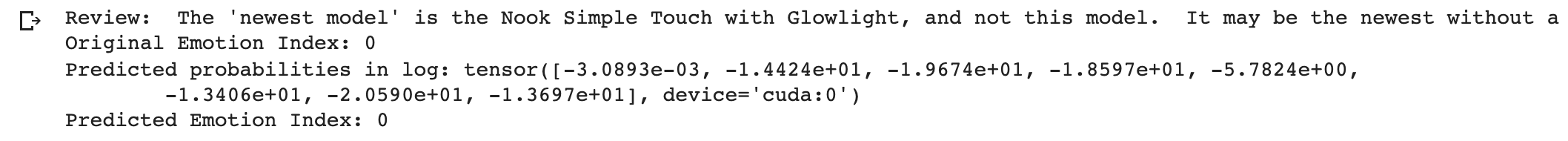}  
    \caption{Results of Emotion Extraction on Sample 2}  
    \label{fig:resultsE2}
\end{figure*}

The model was trained for 100 epochs with the help of SGD optimizer with learning rate of 0.003 and momentum of 0.9. The trained model will be used to calculate the probability of each emotion. Sample results can be seen in Figures \ref{fig:resultsE1} and \ref{fig:resultsE2}.

\subsection{Extraction of Causes}

For finding the cause clause, we create a model which predicts whether a given clause is a cause clause or not. For each of the clause, we first tokenize the clause into words and convert each word into embeddings. Now, we create 8 embeddings for each word - one for each emotion. This is created by multiplying each dimension of embedding with the probability of emotion that is outputted from the emotion extraction model. It’s possible that a review involves multiple emotions and this step takes into account this subtlety. 

The 8 embeddings are passed as an input time step to Bi-LSTM. The Bi-LSTM outputs a 2 * 1024 size vector for each time-step. The output of the last time step of Bi-LSTM is passed through a dropout with probability 0.5. We use the output of the last layer as input to a linear layer. The linear layer outputs a vector of size 80 which is passed through the elu nonlinearity function. The output of this layer is further passed to a linear layer which outputs a vector of size 8. This output is further passed to a sigmoid function which provides the probability of the clause being a cause clause. The training of the model is carried out with the help of the BCE loss function. We are running a stochastic-gradient descent and hence, are using a batch size of 1. The clause having the highest probability is selected as the cause clause.

\begin{figure*}  
    \includegraphics[width=\textwidth]{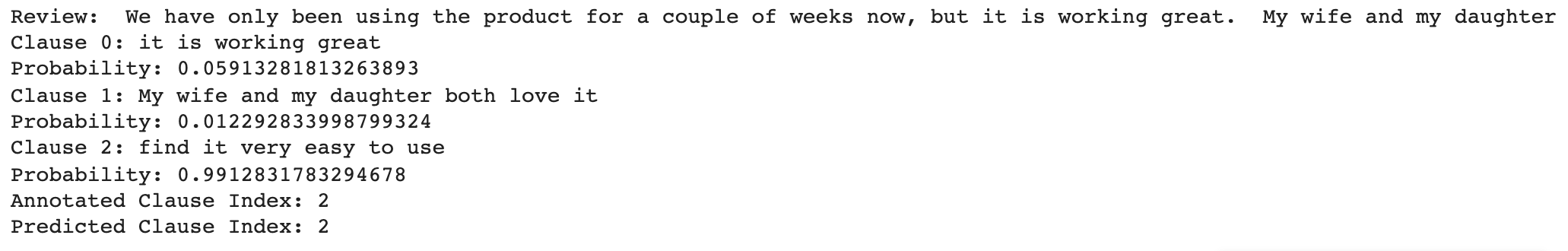}  
    \caption{Results of Cause Extraction on Sample 1}  
    \label{fig:resultsC1}
\end{figure*}

\begin{figure*}  
    \includegraphics[width=\textwidth]{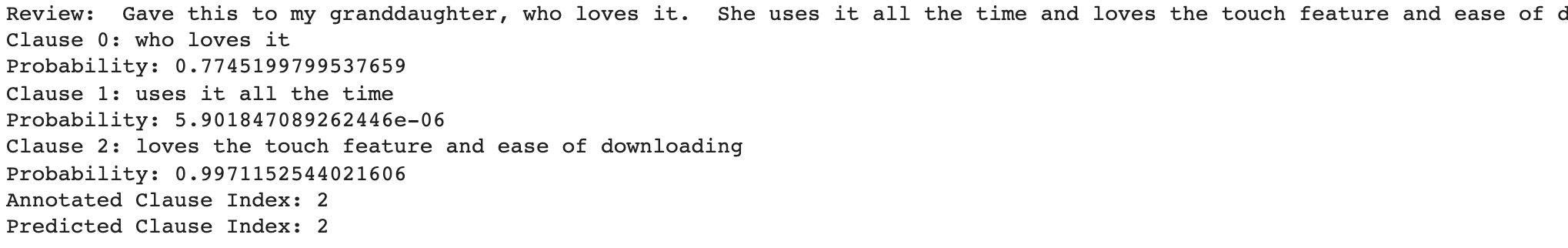}  
    \caption{Results of Cause Extraction on Sample 2}  
    \label{fig:resultsC2}
\end{figure*}

The model was trained for 50 epochs with the help of SGD optimizer with learning rate of 0.003 and momentum of 0.9. Sample results can be seen in Figures \ref{fig:resultsC1} and \ref{fig:resultsC2}.

\subsection{Clustering of Causes}

Once we get a cause clause for each review, we cluster the reviews to cover most concepts. For example: there might be two reviews stating that the product is cheap. Such reviews will be covered in the same cluster and will be represented by one cause. 

To cluster the clauses, we first represent each clause with a vector. Firstly we convert each word of the clause into their corresponding word2vec embedding and emotion-aware embedding. We concatenate these embeddings for each word and then the clause embedding is obtained by max-pooling over these words. Once we get the clause embedding, we use agglomerative clustering to cluster the clauses. We use cosine similarity as the distance function. We are using complete linkage as we want all the clauses in a cluster to be similar to each other. In complete linkage, we calculate the distance between each node of the two clusters and merge the clusters only if the max distance is less than 0.13. We further remove clusters with size < 2 as those clusters are not important.

Once we get a cluster of clauses, we need to represent each cluster with a head clause. The head clause is a clause which is close to every other clause in the cluster. We calculate the largest distance of a clause in a cluster with each clause in a cluster and pick the clause with the smallest max distance as the head clause. 

The clustering is done for each (product, emotion) pair. Hence, we get a list of clauses representing each emotion for the product.

\begin{figure}  
    \includegraphics[scale=0.35]{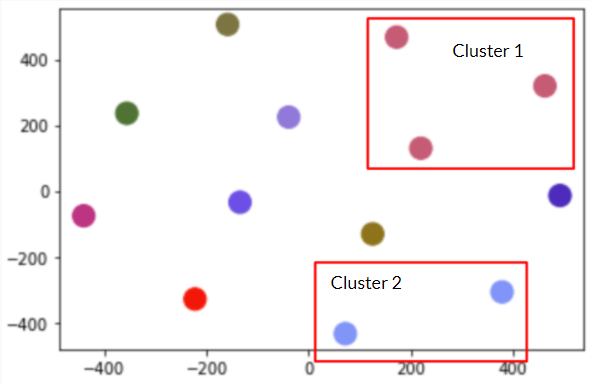}  
    \caption{Clusters of Cause Clauses}  
    \label{fig:clusters}
\end{figure}

\section{Results}

\begin{figure*}  
    \includegraphics[width=\textwidth]{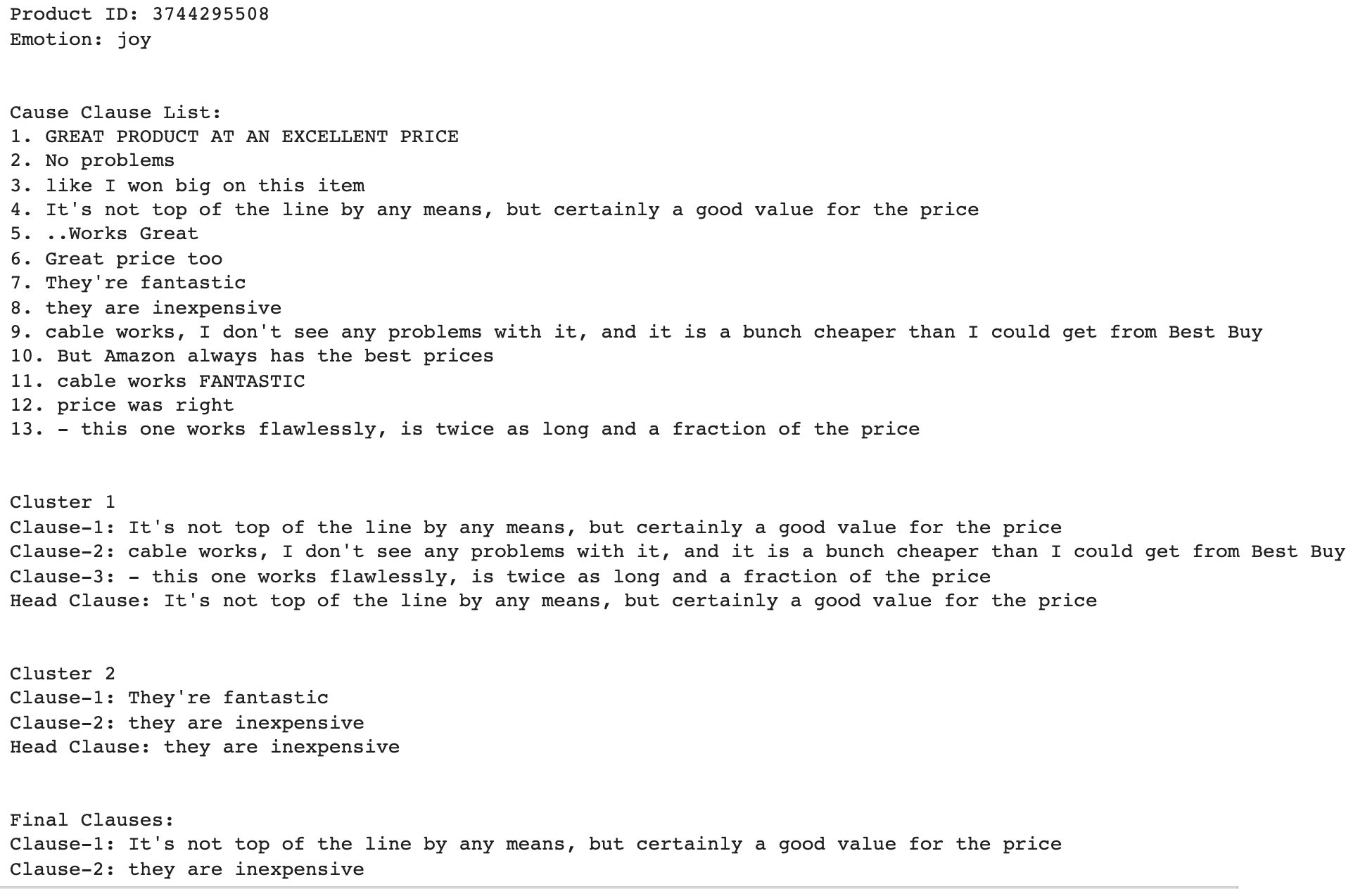}  
    \caption{Results of Emotion Cause Extraction and Clustering}  
    \label{fig:results}
\end{figure*}

The results for a product and emotion can be seen in Figure \ref{fig:results}.

In our dataset, the emotion and cause clause predictions looks to generate some meaningful results.

The affinity of above clauses can be seen in the graph Figure \ref{fig:clusters}. We have used t-SNE algorithm to reduce the vectors to 2-D. Since there are less reviews, the reduction was not that efficient, but it gives an idea about the affinity.

A quick overview of the project in the form of a video can be found here: \url{https://www.youtube.com/watch?v=YEUB32NEZmU}. Access to the code and all related files and resources can be found here: \url{https://github.com/ArpitMittalUSC/Emotion-Cause-Extraction}

\section {Future Work}

The popular field of emotion-related tasks in Natural Language Processing sees no shortage in opportunities to innovate. Although we achieved our initial objective in this project, there is a lot of potential for future work.

Due to time and resource constraints, we have only managed to run this pipeline on a small dataset. However, training and testing this methodology on a much larger scale would leverage the true power of the pipeline.

Another issue was the lack of a benchmark annotated dataset of emotion-cause pairs in the English language. With improved annotations, the results would improve as well, and if a benchmark dataset is created, it would make it much easier to compare our pipeline with other methodologies that exist for ECPE.

Future work can also include using more complex and computationally intensive models such as Transformers and contextual embeddings to improve performance.

\bibliographystyle{acl_natbib}
\bibliography{report}

\begin{thebibliography}{13}
\expandafter\ifx\csname natexlab\endcsname\relax\def\natexlab#1{#1}\fi

\bibitem[{Chen et~al.(2010)Chen, Yat, Lee, Li, and ren
  Huang}]{Chen10emotioncause}
Ying Chen, Sophia Yat, Mei Lee, Shoushan Li, and Chu ren Huang. 2010.
\newblock Emotion cause detection with linguistic constructions.
\newblock In \emph{In: Proceeding of the 23rd International Conference on
  Computational Linguistics (COLING}.

\bibitem[{Cheng et~al.(2020)Cheng, Jiang, Yin, Yu, and
  Gu}]{cheng-etal-2020-symmetric}
Zifeng Cheng, Zhiwei Jiang, Yafeng Yin, Hua Yu, and Qing Gu. 2020.
\newblock \href {https://doi.org/10.18653/v1/2020.coling-main.12} {A symmetric
  local search network for emotion-cause pair extraction}.
\newblock In \emph{Proceedings of the 28th International Conference on
  Computational Linguistics}, pages 139--149, Barcelona, Spain (Online).
  International Committee on Computational Linguistics.

\bibitem[{Gui et~al.(2016)Gui, Wu, Xu, Lu, and Zhou}]{gui-etal-2016-event}
Lin Gui, Dongyin Wu, Ruifeng Xu, Qin Lu, and Yu~Zhou. 2016.
\newblock \href {https://doi.org/10.18653/v1/D16-1170} {Event-driven emotion
  cause extraction with corpus construction}.
\newblock In \emph{Proceedings of the 2016 Conference on Empirical Methods in
  Natural Language Processing}, pages 1639--1649, Austin, Texas. Association
  for Computational Linguistics.

\bibitem[{Gui et~al.(2014)Gui, Yuan, Xu, Liu, Lu, and Zhou}]{GuiYXLLZ14}
Lin Gui, Li~Yuan, Ruifeng Xu, Bin Liu, Qin Lu, and Yu~Zhou. 2014.
\newblock \href {https://doi.org/10.1007/978-3-662-45924-9_42} {Emotion cause
  detection with linguistic construction in chinese weibo text}.
\newblock In \emph{Natural Language Processing and Chinese Computing - Third
  CCF Conference, NLPCC 2014, Shenzhen, China, December 5-9, 2014.
  Proceedings}, volume 496 of \emph{Communications in Computer and Information
  Science}, pages 457--464. Springer.

\bibitem[{Gupta et~al.(2010)Gupta, Gilbert, and
  Di~Fabbrizio}]{gupta-etal-2010-emotion}
Narendra Gupta, Mazin Gilbert, and Giuseppe Di~Fabbrizio. 2010.
\newblock \href {https://aclanthology.org/W10-0202} {Emotion detection in email
  customer care}.
\newblock In \emph{Proceedings of the {NAACL} {HLT} 2010 Workshop on
  Computational Approaches to Analysis and Generation of Emotion in Text},
  pages 10--16, Los Angeles, CA. Association for Computational Linguistics.

\bibitem[{Khunteta and Singh(2021)}]{9478079}
Arunima Khunteta and Pardeep Singh. 2021.
\newblock \href {https://doi.org/10.1109/ICSCCC51823.2021.9478079} {Emotion
  cause extraction - a review of various methods and corpora}.
\newblock In \emph{2021 2nd International Conference on Secure Cyber Computing
  and Communications (ICSCCC)}, pages 314--319.

\bibitem[{Lee et~al.(2010)Lee, Chen, and Huang}]{lee-etal-2010-text}
Sophia Yat~Mei Lee, Ying Chen, and Chu-Ren Huang. 2010.
\newblock \href {https://aclanthology.org/W10-0206} {A text-driven rule-based
  system for emotion cause detection}.
\newblock In \emph{Proceedings of the {NAACL} {HLT} 2010 Workshop on
  Computational Approaches to Analysis and Generation of Emotion in Text},
  pages 45--53, Los Angeles, CA. Association for Computational Linguistics.

\bibitem[{Li et~al.(2018)Li, Song, Feng, Wang, and Zhang}]{li-etal-2018-co}
Xiangju Li, Kaisong Song, Shi Feng, Daling Wang, and Yifei Zhang. 2018.
\newblock \href {https://doi.org/10.18653/v1/D18-1506} {A co-attention neural
  network model for emotion cause analysis with emotional context awareness}.
\newblock In \emph{Proceedings of the 2018 Conference on Empirical Methods in
  Natural Language Processing}, pages 4752--4757, Brussels, Belgium.
  Association for Computational Linguistics.

\bibitem[{Mao et~al.(2019)Mao, Chang, Shi, Li, and Shi}]{app9071334}
Xingliang Mao, Shuai Chang, Jinjing Shi, Fangfang Li, and Ronghua Shi. 2019.
\newblock \href {https://doi.org/10.3390/app9071334} {Sentiment-aware word
  embedding for emotion classification}.
\newblock \emph{Applied Sciences}, 9(7).

\bibitem[{Russo et~al.(2011)Russo, Caselli, Rubino, Boldrini, and
  Mart{\'\i}nez-Barco}]{russo-etal-2011-emocause}
Irene Russo, Tommaso Caselli, Francesco Rubino, Ester Boldrini, and Patricio
  Mart{\'\i}nez-Barco. 2011.
\newblock \href {https://aclanthology.org/W11-1720} {{EMOC}ause: An
  easy-adaptable approach to extract emotion cause contexts}.
\newblock In \emph{Proceedings of the 2nd Workshop on Computational Approaches
  to Subjectivity and Sentiment Analysis ({WASSA} 2.011)}, pages 153--160,
  Portland, Oregon. Association for Computational Linguistics.

\bibitem[{Singh et~al.(2021)Singh, Hingane, Wani, and Modi}]{singh2021endtoend}
Aaditya Singh, Shreeshail Hingane, Saim Wani, and Ashutosh Modi. 2021.
\newblock \href {http://arxiv.org/abs/2103.01544} {An end-to-end network for
  emotion-cause pair extraction}.

\bibitem[{Xia and Ding(2019)}]{xia-ding-2019-emotion}
Rui Xia and Zixiang Ding. 2019.
\newblock \href {https://doi.org/10.18653/v1/P19-1096} {Emotion-cause pair
  extraction: A new task to emotion analysis in texts}.
\newblock In \emph{Proceedings of the 57th Annual Meeting of the Association
  for Computational Linguistics}, pages 1003--1012, Florence, Italy.
  Association for Computational Linguistics.

\bibitem[{Xu et~al.(2017)Xu, Hu, Lu, Wu, and Gui}]{8195347}
Ruifeng Xu, Jiannan Hu, Qin Lu, Dongyin Wu, and Lin Gui. 2017.
\newblock \href {https://doi.org/10.23919/TST.2017.8195347} {An ensemble
  approach for emotion cause detection with event extraction and multi-kernel
  svms}.
\newblock \emph{Tsinghua Science and Technology}, 22(6):646--659.

\end{thebibliography}


\end{document}